\documentclass{article}

\usepackage[final]{neurips_2025}

\usepackage[utf8]{inputenc} 
\usepackage[T1]{fontenc}    
\usepackage{hyperref}       
\usepackage{url}            
\usepackage{booktabs}       
\usepackage{amsfonts}       
\usepackage{nicefrac}       
\usepackage{microtype}      
\usepackage{xcolor}         
\usepackage{multirow}
\usepackage{graphicx}
\usepackage{amsmath}
\usepackage{listings}
\usepackage{xcolor}
\usepackage{comment}

\definecolor{myblue}{RGB}{0,0,180}

\title{Do LLMs Benefit from User and Item Embeddings in Recommendation Tasks?}

\author{%
  Mir Rayat Imtiaz Hossain\thanks{Work done during an internship at RBC Borealis.} \\
  University of British Columbia\\
  \And
  Leo Feng\thanks{Correspondence to: Leo Feng \href{mailto:leo.feng@borealisai.com}{<leo.feng@borealisai.com>}}\\
  RBC Borealis \\
  \AND
  Leonid Sigal \\
  University of British Columbia \\
  \And
  Mohamed Osama Ahmed \\
  RBC Borealis \\
}

\begin{document}

\maketitle

\begin{abstract}
Large Language Models (LLMs) have emerged as promising recommendation systems, offering novel ways to model user preferences through generative approaches. However, many existing methods often rely solely on text semantics or incorporate collaborative signals in a limited manner, typically using only user or item embeddings. These methods struggle to handle multiple item embeddings representing user history, reverting to textual semantics and neglecting richer collaborative information. In this work, we propose a simple yet effective solution that projects user and item embeddings, learned from collaborative filtering, into the LLM token space via separate lightweight projector modules. A finetuned LLM then conditions on these projected embeddings alongside textual tokens to generate recommendations. Preliminary results show that this design effectively leverages structured user–item interaction data, improves recommendation performance over text-only LLM baselines, and offers a practical path for bridging traditional recommendation systems with modern LLMs.
\end{abstract}

\section{Introduction}

Large Language Models (LLMs) have grown into widely adopted tool across a broad range of natural language processing~\cite{brown2020language,touvron2023llama} and multi-modal tasks. Such tasks include captioning and question answering with images~\cite{alayrac2022flamingo,li2023blip, liu2023visual, team2024qwen2}, videos~\cite{hui2024qwen2, li2024llava} and other types of embeddings~\cite{hong20233d, tennenholtz2024demystifying};  demonstrating strong generalization, reasoning and language generation capabilities. Due to the remarkable capabilities of the LLMs, they have also been widely adopted for recommendation systems~\cite{bao2023tallrec, cui2022m6, geng2022recommendation, lyu2024llm, wei2024llmrec, xu2024openp5}. However, these approaches rely purely on text modality and hence fall short of capturing the rich user-item interactions and item co-occurence relationships.

Inspired by the multi-modal language models~\cite{alayrac2022flamingo, hui2024qwen2, li2024llava, li2023blip, liu2023visual}, several recent recommendation system approaches~\cite{wang2025enhancing, yang2024item, zhang2024text, zhang2025collm} proposed injecting user and item latent embeddings learned from collaborative filters. However, these works~\cite{wang2025enhancing, zhang2024text, zhang2025collm} are limited to binary classification tasks and can only handle a single target item embedding at a time. ILM~\cite{yang2024item}, in contrast, focuses on item embeddings, ignoring user embeddings.

Our approach explores injecting user and item collaborative embeddings into the LLMs so that the LLMs can learn and capture rich user-item interaction patterns and co-occurrence relationships, differing from existing work in several directions:

(1) Instead of conditioning on a single embedding, our approach allows the LLM to take as input a user embedding along with an arbitrary number of item embeddings, capturing a richer representation of the user’s history by jointly leveraging both user- and item-level information.

(2) We introduce two separate lightweight projectors for users and items, mapping them into the language space independently allowing the LLM to learn complementary representations.

Our results show that this design significantly improves recommendation performance over text-only LLM baselines and narrows the gap with classical recommendation systems, even surpassing them in certain scenarios. These findings highlight the promise of embedding-grounded LLMs as a practical path toward bridging traditional collaborative filtering with modern generative recommendation.

\vspace{-0.15in}

\section{Method}
\vspace{-0.1in}
\begin{figure}[t]
\centering
\includegraphics[width=0.75\textwidth]{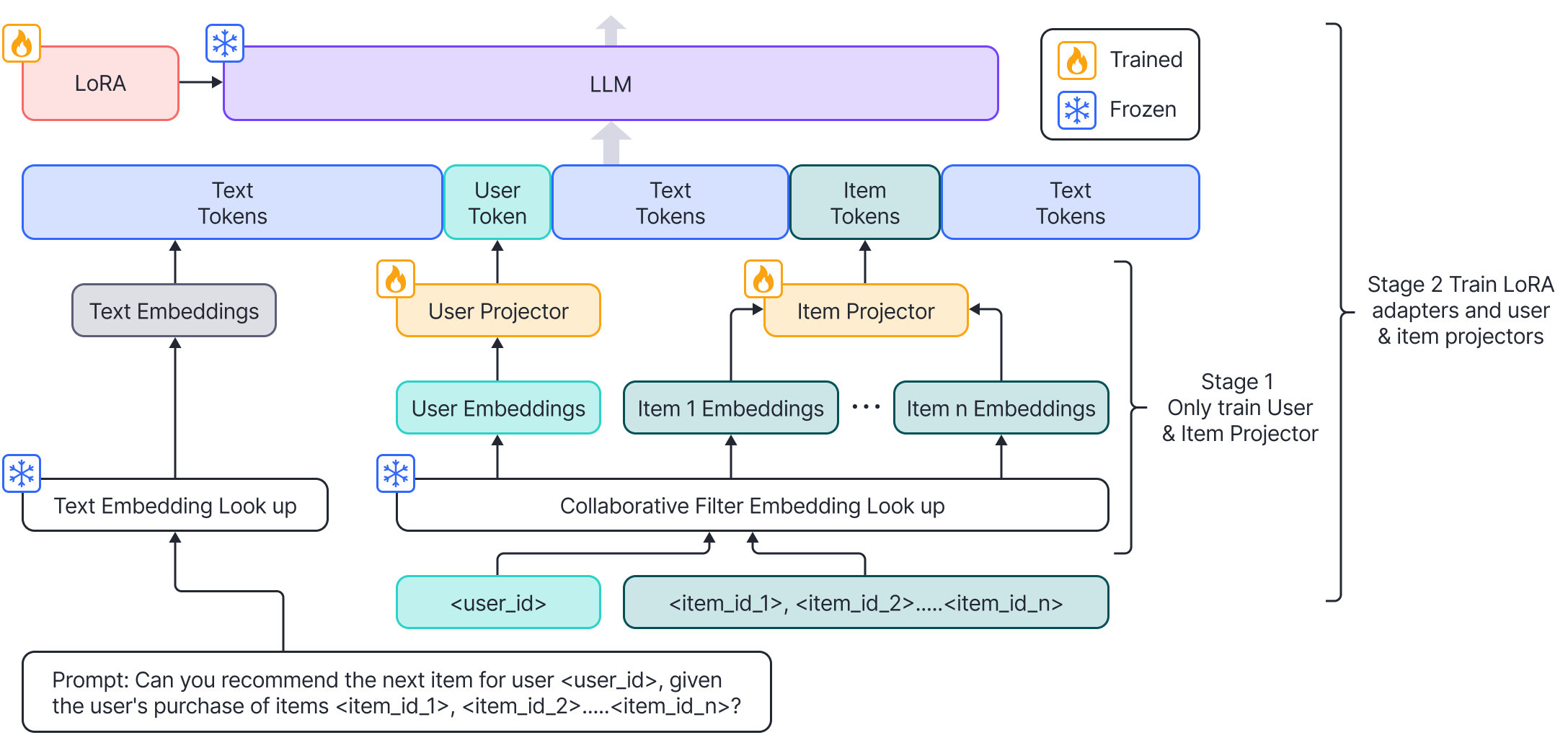}
\caption{{\bf Overview of our proposed framework.} User and item embeddings from collaborative filtering are projected into the language space and injected into the LLM alongside textual tokens. Training proceeds in two stages (see Section~\ref{subsec:stages}).}
\vspace{-0.25in}
\label{fig:framework}
\end{figure}

LLMs are inherently text-centric and therefore struggle to capture the rich user–item interaction patterns and co-occurrence structures that collaborative filtering methods naturally exploit. To address this, we inject collaborative filtering embeddings for both users and items into the LLM. Our design (Figure~\ref{subsec:stages}) allows the model to condition on a user embedding together with an arbitrary number of item embeddings from the user’s history, providing richer contextual signals than text-only inputs. 

\vspace{-0.10in}

\subsection{Recommendation Task and Prompt Template}

In natural language, a common way to request a recommendation is to mention the user and their interaction history. For example, a prompt \cite{xu2024openp5} might look like this:


\vspace{-0.02in}


\begin{lstlisting}
Input: Considering (*@\textcolor{myblue}{\{dataset\}}@*) user (*@\textcolor{myblue}{\{user\_id\}}@*) has interacted with (*@\textcolor{myblue}{\{dataset\}}@*) items (*@\textcolor{myblue}{\{history\}}@*). What is the next recommendation for the user?
Target: (*@\textcolor{myblue}{\{dataset\}}@*) (*@\textcolor{myblue}{\{target\}}@*)
\end{lstlisting}

\vspace{-0.10in}
In this template, \texttt{\{dataset\}} denotes the dataset name, \texttt{\{user\_id\}} the user identifier, \texttt{\{history\}} the list of items the user has interacted with, and \texttt{\{target\}} the ground-truth next item.

\vspace{-0.10in}
\subsection{Model Architecture} 

Tackling this setting, our framework, illustrated in Figure~\ref{fig:framework},  consists of five components:  
(i) a text tokenizer;  
(ii) a collaborative filter lookup table;  
(iii) a user embedding projector;  
(iv) an item embedding projector;  
(v) the LLM backbone.  

First, we pass the text prompt to LLM tokenizer to extract the text embeddings $emb(t_k)$ of each text token $t_k$. However, since we do not use textual representations for \texttt{\{user\_id\}} and \texttt{\{item\_id\}} for items in the \texttt{\{history\}}, we replace textual embeddings with pre-computed collaborative filtering embeddings retrieved from a lookup table.

To learn these embeddings, we perform matrix factorization on the user–item interaction data using \textit{Weighted Alternating Least Squares (WALS)}\footnote{More sophisticated collaborative filtering techniques could be applied. For this work, we consider matrix factorization as it offers a simple yet effective foundation for our framework.}~\cite{hu2008collaborative}, where the dot product between user embeddings $u$ and item embeddings $i$ approximates the preference score. 

Since LLMs expect language-space representations, we introduce two separate projectors: a user projector $f_u(u)$ and an item projector $f_i(i)$. Both are two-layer MLPs that map collaborative embeddings into the LLM token space:
\begin{equation}
\begin{aligned}
emb(u) &= f_u(u), \\
emb(i) &= f_i(i).
\end{aligned}
\end{equation}

The resulting projected embeddings are injected into the LLM alongside text embeddings. Training is performed with a standard next-token prediction loss.

\subsection{Training Stages} 
\label{subsec:stages}

Because our framework introduces projector modules to project the matrix factorization embeddings of user and items to the language space, we employ a two-stage fine-tuning strategy:  

\noindent
\textbf{Stage 1 — Projector pre-training.} We freeze the LLM and fine-tune only the user and item projectors, enabling them to learn meaningful mappings into the language space. 

\noindent
\textbf{Stage 2 — Joint fine-tuning.} Once the projectors are initialized, we fine-tune them jointly with LoRA adapters on the LLM backbone. This allows the model to adapt both the projected embeddings and the language model parameters for improved recommendation performance.

\section{Experiments}

\subsection{Experimental Setting}

\textbf{Datasets.}
We evaluate our framework on three datasets from the OpenP5 library: \textbf{Amazon Beauty}, \textbf{LastFM}, and \textbf{MovieLens-1M}. Each dataset contains user–item interaction histories and has two recommendation tasks: \textbf{Sequential} and \textbf{Straightforward} Recommendation. In the Sequential setting, the model uses user information and item interaction history to predict the next item. In the Straightforward setting, the model predicts the next item given only the user ID.

\textbf{Prompt Construction.} For both recommendation tasks, we follow OpenP5~\cite{xu2024openp5}, which provides 11 prompt templates per task. As in~\cite{xu2024openp5}, we train on 10 templates and reserve the remaining one for zero-shot generalization (\texttt{unseen} setting).

\textbf{Architecture.} Our goal is to assess how collaborative filtering embeddings enhance GPT-style decoder-only LLMs~\cite{brown2020language}. Following the OpenP5 library~\cite{xu2024openp5}, we use the OpenLLaMA-3B model, a reproduction of LLaMA-2~\cite{touvron2023llama}.

\textbf{Training Process.} Following OpenP5~\cite{xu2024openp5}, we alternate between the two tasks during training. In the Straightforward case, only the user projector is active since no item history is provided.

\textbf{Metrics.} We evaluate model performance using two standard metrics in recommendation systems: top-$k$ Hit Ratio (HR@$k$) and Normalized Discounted Cumulative Gain (NDCG@$k$), reported at $k = 5$ and $k = 10$. HR@$k$ checks if the ground-truth item appears in the top-$k$ ranked list, while NDCG@$k$ considers the position of the correct item, favoring higher scores for items ranked closer to the top. These metrics jointly capture both accuracy and ranking quality.

\subsection{Results}

Tables~\ref{tab:seqrec_clean} and \ref{tab:straightforward_llama_only} show results for sequential and straightforward recommendation tasks. Decoder-only LLMs without collaborative embeddings (OpenP5's Llama-R/S/C) perform poorly, highlighting the limitations of text-only approaches. In contrast, our embedding-augmented model (Llama-Embed-Stage-2) achieves significant gains across all datasets. On the sequential task (Table~\ref{tab:seqrec_clean}), it bridges the gap between text-only LLMs with that of strong traditional recommendation systems like SASRec and HGN, surpassing their performance on Amazon Beauty, demonstrating the effectiveness of collaborative embeddings that leverage user–item structure.

\begin{table}[t]
\centering
\small
\setlength{\tabcolsep}{2.5pt}
\resizebox{\textwidth}{!}{%
\begin{tabular}{lcccc|cccc|cccc}
\toprule
\multirow{2}{*}{Methods} &
\multicolumn{4}{c|}{Beauty} &
\multicolumn{4}{c|}{LastFM} &
\multicolumn{4}{c}{ML1M} \\
& HR@5 & NDCG@5 & HR@10 & NDCG@10
& HR@5 & NDCG@5 & HR@10 & NDCG@10
& HR@5 & NDCG@5 & HR@10 & NDCG@10 \\
\midrule

\multicolumn{13}{l}{\textit{Traditional Recommender Systems}} \\ 

Caser~\cite{tang2018personalized}   & 0.0205 & 0.0131 & 0.0347 & 0.0176
        & 0.0303 & 0.0178 & 0.0413 & 0.0214
        & 0.0912 & 0.0565 & 0.1442 & 0.0734 \\
HGN~\cite{ma2019hierarchical}     & 0.0325 & 0.0206 & 0.0512 & 0.0266
        & 0.0321 & 0.0175 & 0.0505 & 0.0233
        & \underline{\textbf{0.1430}} & \underline{\textbf{0.0874}} & \underline{\textbf{0.2404}} & \underline{\textbf{0.1231}} \\
GRU4Rec~\cite{hidasi2015session} & 0.0164 & 0.0099 & 0.0283 & 0.0137
        & 0.0275 & 0.0158 & 0.0367 & 0.0187
        & 0.0806 & 0.0475 & 0.1344 & 0.0649 \\
BERT4Rec~\cite{sun2019bert4rec} & 0.0203 & 0.0124 & 0.0347 & 0.0170
        & 0.0422 & 0.0269 & 0.0633 & 0.0337
        & 0.1308 & 0.0804 & 0.2219 & 0.1097 \\
FDSA~\cite{zhang2019feature}   & 0.0267 & 0.0163 & 0.0407 & 0.0208
        & 0.0303 & 0.0219 & 0.0413 & 0.0254
        & 0.1167 & 0.0762 & 0.1868 & 0.0987 \\
SASRec~\cite{kang2018self}  & \textbf{0.0387} & \textbf{0.0249} & \textbf{0.0605} & \textbf{0.0318}
        & \underline{\textbf{0.0505}} & \underline{\textbf{0.0331}} & \underline{\textbf{0.0688}} & \underline{\textbf{0.0390}}
        & 0.1078 & 0.0681 & 0.1810 & 0.0918 \\
\midrule

\multicolumn{13}{l}{\textit{LLM-based Methods (Seen Prompts)}} \\
ILM~\cite{yang2024item} & 0.0213 & 0.0164 & 0.0270 & 0.0182
        & - & - & - & -
        & 0.0724 & 0.0485  & 0.1064 & 0.0595 \\
Llama-R~\cite{xu2024openp5} & 0.0018 & 0.0013 & 0.0024 & 0.0015
        & 0.0193 & 0.0120 & 0.0284 & 0.0149
        & 0.0300 & 0.0197 & 0.0470 & 0.0252 \\
Llama-S~\cite{xu2024openp5} & 0.0022 & 0.0036 & 0.0013 & 0.0017
        & 0.0101 & 0.0059 & 0.0202 & 0.0092
        & 0.0714 & 0.0466 & 0.1094 & 0.0587 \\
Llama-C~\cite{xu2024openp5} & 0.0002 & 0.0001 & 0.0007 & 0.0003
        & 0.0018 & 0.0013 & 0.0018 & 0.0013
        & 0.0012 & 0.0006 & 0.0026 & 0.0011 \\
Llama-Embed-Stage-1 (Ours) & 0.0361 & 0.0318 & 0.0421 & 0.0337
        & 0.0294 & 0.0172 & 0.0468 & 0.0227
        & 0.0679 & 0.0459 & 0.1026 & 0.0570 \\
Llama-Embed-Stage-2 (Ours) & \underline{\textbf{0.0642}} & \underline{\textbf{0.0514}} & \underline{\textbf{0.0794}} & \underline{\textbf{0.0563}}
        & \textbf{0.0422} & \textbf{0.0258} & \textbf{0.0615} & \textbf{0.0322}
        & \textbf{0.1109} &\textbf{ 0.0726} & \textbf{0.1786} & \textbf{0.0943} \\
\midrule

\multicolumn{13}{l}{\textit{LLM-based Methods (Unseen Prompts)}} \\
ILM~\cite{yang2024item}  & 0.0213 & 0.0162 & 0.0269 & 0.0181
        & - & - & - & -
        & 0.0717 & 0.0481 & 0.1086 & 0.0600 \\
Llama-R~\cite{xu2024openp5}  & 0.0017 & 0.0011 & 0.0022 & 0.0013
        & 0.0183 & 0.0108 & 0.0202 & 0.0113
        & 0.0296 & 0.0200 & 0.0444 & 0.0247 \\
Llama-S~\cite{xu2024openp5}  & 0.0029 & 0.0017 & 0.0045 & 0.0022
        & 0.0128 & 0.0078 & 0.0202 & 0.0103
        & 0.0556 & 0.0364 & 0.0877 & 0.0467 \\
Llama-C~\cite{xu2024openp5}  & 0.0004 & 0.0002 & 0.0007 & 0.0003
        & 0.0009 & 0.0004 & 0.0046 & 0.0015
        & 0.0010 & 0.0006 & 0.0018 & 0.0009 \\
Llama-Embed-Stage-1 (Ours) & 0.0333 & 0.0278 & 0.0393 & 0.0297
        & 0.0248 & 0.0162 & 0.0395 & 0.0210
        & 0.0666 & 0.0445 & 0.1028 & 0.0561 \\
Llama-Embed-Stage-2 (Ours) & \textbf{0.0631} & \textbf{0.0507} & \textbf{0.0791} & \textbf{0.0559}
        & \textbf{0.0413} & \textbf{0.0258} & \textbf{0.0624} & \textbf{0.0382}
        & \textbf{0.1132} & \textbf{0.0731} & \textbf{0.1790} &\textbf{0.0941} \\
\bottomrule
\end{tabular}%
}
\vspace{2pt}
\caption{Performance on the \textbf{sequential} recommendation task. The Llama-R, Llama-S, and Llama-C correspond to OpenP5's trained models leveraging their \texttt{random}, \texttt{sequential}, and \texttt{collaborative} indexing respectively. Bold numbers indicate the best results within each group, 
while bold + underlined numbers indicate the best results overall across all groups.}
\label{tab:seqrec_clean}
\vspace{-0.1in}
\end{table}

\begin{table}[t]
\centering
\small
\setlength{\tabcolsep}{4.5pt}
\resizebox{\textwidth}{!}{%
\begin{tabular}{lcccc|cccc|cccc}
\toprule
\multirow{2}{*}{Methods} &
\multicolumn{4}{c|}{Beauty} &
\multicolumn{4}{c|}{LastFM} &
\multicolumn{4}{c}{ML1M} \\
& HR@5 & NDCG@5 & HR@10 & NDCG@10
& HR@5 & NDCG@5 & HR@10 & NDCG@10
& HR@5 & NDCG@5 & HR@10 & NDCG@10 \\
\midrule

\multicolumn{13}{l}{\textit{Traditional Recommender Systems}} \\
BPR\text{-}MF~\cite{rendle2012bpr}   & 0.0224 & 0.0149 & 0.0363 & 0.0204
                & 0.0218 & 0.0147 & 0.0253 & 0.0162
                & 0.0141 & 0.0081 & 0.0301 & 0.0133 \\
BPR\text{-}MLP~\cite{cheng2016wide}  & 0.0193 & 0.0127 & 0.0305 & 0.0176
                & 0.0211 & 0.0150 & 0.0321 & 0.0185
                & 0.0123 & 0.0068 & 0.0270 & 0.0116 \\
SimpleX~\cite{mao2021simplex} & \textbf{0.0300} & \textbf{0.0189} & \textbf{0.0471} & \textbf{0.0245}
                & \textbf{0.0312} & \textbf{0.0211} &\textbf{ 0.0523 }& \textbf{0.0277}
                & \underline{\textbf{0.0301}} & \underline{\textbf{0.0133}} & \underline{\textbf{0.0596}} & \underline{\textbf{0.0206}} \\
\midrule

\multicolumn{13}{l}{\textit{LLM-based Methods (Seen Prompts)}} \\
Llama\text{-}R~\cite{xu2024openp5} 
                & 0.0014 & 0.0021 & 0.0011 & 0.0013
                & 0.0122 & 0.0122 & 0.0275 & 0.0146
                & 0.0106 & 0.0061 & 0.0205 & 0.0093 \\
Llama\text{-}S~\cite{xu2024openp5} 
                & 0.0050 & 0.0035 & 0.0065 & 0.0040
                & 0.0147 & 0.0112 & 0.0220 & 0.0134
                & 0.0103 & 0.0066 & 0.0210 & 0.0104 \\
Llama\text{-}C~\cite{xu2024openp5} 
                & 0.0003 & 0.0001 & 0.0004 & 0.0002
                & 0.0028 & 0.0046 & 0.0022 & 0.0028
                & 0.0012 & 0.0007 & 0.0022 & 0.0010 \\
Llama-Embed-Stage-1 (Ours) &  0.0315 & 0.0241 & 0.0370 & 0.0259
        &   0.0330 & 0.0195 & 0.0459 & 0.0235 
        &    0.0179 & 0.0103 & 0.0301 & 0.0142 \\
Llama-Embed-Stage-2 (Ours) &   \textbf{0.0586} & \textbf{0.0467} & \textbf{0.0730} & \textbf{0.0513}  & 
\textbf{0.0505} & \textbf{0.0330} & \textbf{0.0697} & \textbf{0.0391} 
&  \textbf{0.0205} & \textbf{0.0117} & \textbf{0.0419} & \textbf{0.0185} \\
\midrule

\multicolumn{13}{l}{\textit{LLM-based Methods (Unseen Prompts)}} \\
Llama\text{-}R~\cite{xu2024openp5} 
                & 0.0011 & 0.0013 & 0.0020 & 0.0013
                & 0.0121 & 0.0103 & 0.0202 & 0.0139
                & 0.0094 & 0.0063 & 0.0190 & 0.0094 \\
Llama\text{-}S~\cite{xu2024openp5} 
                & 0.0047 & 0.0032 & 0.0062 & 0.0038
                & 0.0147 & 0.0108 & 0.0202 & 0.0126
                & 0.0098 & 0.0066 & 0.0195 & 0.0097 \\
Llama\text{-}C~\cite{xu2024openp5} 
                & 0.0004 & 0.0002 & 0.0004 & 0.0002
                & 0.0037 & 0.0028 & 0.0037 & 0.0028
                & 0.0005 & 0.0003 & 0.0015 & 0.0006 \\
Llama-Embed-Stage-1 (Ours) &  0.0334 & 0.0266 & 0.0389 & 0.0284
        &  0.0284 & 0.0165 & 0.0385 & 0.0198
        & 0.0189 & 0.0113 & 0.0313 & 0.0153 \\
Llama-Embed-Stage-2 (Ours) &  \underline{\textbf{0.0593}} & \underline{\textbf{0.0473}} & \underline{\textbf{0.0744}} & \underline{\textbf{0.0521}}
        &  \underline{\textbf{0.0514 }}& \underline{\textbf{0.0344}} & \underline{\textbf{0.0725}} & \underline{\textbf{0.0412}}
        & \textbf{0.0199} & \textbf{0.0114} & \textbf{0.0412} & \textbf{0.0182} \\
\bottomrule
\end{tabular}%
}
\vspace{2pt}
\caption{Performance on the \textbf{straightforward} recommendation task. }
\vspace{-0.35in}
\label{tab:straightforward_llama_only}
\end{table}

\textbf{Stage 1 vs. Stage 2.} Consistent and substantial gains can be seen when moving from Stage 1 (projector-only fine-tuning) to Stage 2 (joint optimization of LoRA adapters and projectors). In Stage 1, the individual projectors learn to project the user and item embeddings to language space but the individual LLM parameters are not tuned to accommodate for the injected embeddings.

\textbf{User-embeddings alone.} On the straightforward task (Table~\ref{tab:straightforward_llama_only}), where only the user is provided, our method achieves strong performance. In general, traditional recommender systems' performance dropped significantly across all 3 datasets without access to item history. In contrast, our collaborative embedding-conditioned LLM approach is more robust, retaining good performance on 2/3 datasets (Beauty and LastFM), outperforming traditional recommender systems.

\textbf{Seen vs unseen prompt settings.} In both Tables \ref{tab:seqrec_clean} and \ref{tab:straightforward_llama_only}, text-only Llama variants introduced in \cite{xu2024openp5} show poor generalization on unseen templates, while our model maintains strong performance, highlighting the robustness achieved by grounding the model with collaborative embeddings.

\section{Conclusion and Future Directions}

In this work, we showed that decoder-only LLMs perform poorly on recommendation tasks when limited to text-only inputs, often resulting in poor performance regardless of the type of indexing used for items. By injecting the model with collaborative filtering embeddings, our framework achieved significant gains across benchmarks, closing the gap and in some cases surpassing the performance of classical recommendation systems.

For future directions, encoding richer item representations (e.g., semantic embeddings from product descriptions), more advanced recommendation system architectures, and improved methods for encoding user–item interactions could further enhance LLM-based recommendation systems. We hope these directions inspire continued research at the intersection of LLMs and recommendation.

\bibliography{main}
\bibliographystyle{plain} 

\end{document}